\documentclass[11pt,a4paper]{article}
\usepackage[hyperref]{acl2019} 
\usepackage{times}
\usepackage{latexsym}

\usepackage{url}

\usepackage{graphicx}
\usepackage{amsmath}
\usepackage{amssymb}
\usepackage{booktabs}
\usepackage{tabulary}
\usepackage{placeins}

\aclfinalcopy 

\title{Evaluating Recurrent Neural Network Explanations}

\author{Leila Arras$^1$, Ahmed Osman$^1$, Klaus-Robert M{\"u}ller$^{2,3,4}$, and Wojciech Samek$^1$\\
  $^1$Machine Learning Group, Fraunhofer Heinrich Hertz Institute, Berlin, Germany\\
  $^2$Machine Learning Group, Technische Universit\"at Berlin, Berlin, Germany\\
  $^3$Department of Brain and Cognitive Engineering, Korea University, Seoul, Korea\\
  $^4$Max Planck Institute for Informatics, Saarbr{\"u}cken, Germany\\
  {\tt \{leila.arras, wojciech.samek\}@hhi.fraunhofer.de}}  

\date{}

\begin{document}
\maketitle
\begin{abstract}
Recently, several methods have been proposed to explain the predictions of recurrent neural networks (RNNs), in particular of LSTMs.
The goal of these methods is to understand the network's decisions by assigning to each input variable, e.g., a word, a {\it relevance} indicating to which extent it contributed to a particular prediction. 
In previous works, some of these methods were not yet compared to one another, or were evaluated only qualitatively.
We close this gap by systematically and quantitatively comparing these methods in different settings, namely (1) a toy arithmetic task which we use as a sanity check, (2) a five-class sentiment prediction of movie reviews, and besides (3) we explore the usefulness of word relevances to build sentence-level representations.
Lastly, using the method that performed best in our experiments, we show how specific linguistic phenomena such as the negation in sentiment analysis reflect in terms of relevance patterns, and how the relevance visualization can help to understand the misclassification of individual samples.
\end{abstract}

\section{Introduction}

Recurrent neural networks such as LSTMs \cite{Hochreiter:1997} are a standard building block for understanding and generating text data in NLP. 
They find usage in pure NLP applications, such as abstractive summarization \cite{Chopra:2016}, machine translation \cite{Bahdanau:ICLR2015}, textual entailment \cite{Rock:ICLR2016}; as well as in multi-modal tasks involving NLP, such as image captioning \cite{Karpathy:CVPR2015}, visual question answering \cite{Xu:ECCV2016} or lip reading \cite{Chung:CVPR2017}. 

As these models become more and more widespread due to their predictive performance, there is also a need to understand {\it why} they took a particular decision, i.e., when the input is a sequence of words: {\it which} words are determinant for the final decision?
This information is crucial to unmask ``Clever Hans'' predictors \cite{Lap:Nature19}, and to allow for transparency of the decision-making process \cite{EU-GDPR}.

Early works on {\it explaining} neural network predictions include \citet{Baehrens:JMLR2010,Zeiler:ECCV2014,Simonyan:ICLR2014,Spring:ICLR2015,Bach:PLOS2015,Alain:ICLR2017}, with several works focusing on explaining the decisions of convolutional neural networks (CNNs) for image recognition. 
More recently, this topic found a growing interest within NLP, amongst others to explain the decisions of general CNN classifiers \cite{Arras:PLOSONE2017,jacovi:BlackboxNLP2018}, and more  particularly to explain the predictions of recurrent neural networks \cite{Li:NAACL2016,Li:ArXiv2017,Arras:WASSA2017,Ding:ACL2017,Murdoch:ICLR2018,Poerner:ACL2018}.

In this work, we focus on RNN explanation methods that are solely based on a trained neural network model and a single test data point\footnote{These methods are deterministic, and are essentially based on a {\it decomposition} of the model's current prediction. Thereby they intend to reflect the {\it sole} model's ``point of view'' on the test data point, and hence are not meant to provide an averaged, smoothed or denoised explanation of the prediction by additionally exploiting the data's {distribution}.}. 
Thus, methods that use additional information, such as training data statistics, sampling, or are optimization-based \cite{Ribeiro:KDD2016,Lundberg:NIPS2017,Chen:ICML2018} are out of our scope.
Among the methods we consider, 
we note that the method of \citet{Murdoch:ICLR2018} was not yet compared against \citet{Arras:WASSA2017,Ding:ACL2017}; and that the method of \citet{Ding:ACL2017} was validated only visually.
Moreover, to the best of our knowledge, no recurrent neural network explanation method was tested so far on a toy problem where the {\it ground truth} relevance value is known.

Therefore our contributions are the following: we evaluate and compare the aforementioned methods, using two different experimental setups, thereby we assess basic properties and  differences between the explanation methods.
Along-the-way we purposely adapted a simple toy task, to serve as a testbed for recurrent neural networks explanations.
Lastly, we explore how word relevances can be used to build sentence-level representations, and demonstrate how the relevance visualization can help to understand the (mis-)classification of selected samples w.r.t.\ semantic composition.

\section{Explaining Recurrent Neural Network Predictions}\label{section:methods}

First, let us settle some notations.
We suppose given a trained recurrent neural network based model, which has learned some scalar-valued prediction function $f_c(\cdot)$, for each class $c$ of a classification problem.
Further, we denote by $\mathbf{x} = (\boldsymbol{x_1}, \boldsymbol{x_2},...,\boldsymbol{x_T})$ an unseen input data point,
where $\boldsymbol{x_t}$ represents the $t$-th input vector of dimension $D$, within the input sequence $\mathbf{x}$ of length $T$.
In NLP, the vectors $\boldsymbol{x_t}$ are typically word embeddings, and $\mathbf{x}$ may be a sentence.

Now, we are interested in methods that can explain the network's prediction $f_c(\mathbf{x})$ for the input $\mathbf{x}$, and for a chosen {\it target} class $c$, by assigning a scalar {relevance} value
to each input variable or word. This relevance is meant to quantify the variable's or word's importance {\it for} or {\it against} a model's prediction {\it towards} the class $c$. 
We denote by $R_{x_i}$ (index $i$) the relevance of a {\it single} variable. This means $x_i$ stands for any arbitrary input variable $x_{t,d}$ representing the $d$-th dimension, $d \in \{1,...,D\}$, of an input vector $\boldsymbol{x_t}$.
Further, we refer to $R_{\boldsymbol{x_t}}$ (index $t$) to designate the relevance value of an {\it entire} input vector or word $\boldsymbol{x_t}$.
Note that, for most methods, one can obtain a word-level relevance value by simply summing up the relevances over the word embedding dimensions, i.e.\ $R_{\boldsymbol{x_t}}= \sum_{d \in \{1,...,D\}}{R_{x_{t,d}}}$\,.

\subsection{Gradient-based explanation}
One standard approach to obtain relevances is based on partial derivatives of the prediction function:  
$R_{x_i} = \big| \tfrac{\partial {f_c}}{\partial x_{i}}(\mathbf{x}) \big|$, or $R_{x_i} = \big( \tfrac{\partial {f_c}}{\partial x_{i}}(\mathbf{x}) \big)^2$ \cite{Dimopoulos:1995,Gevrey:2003,Simonyan:ICLR2014,Li:NAACL2016}.

In NLP this technique was employed to visualize the relevance of single input variables in RNNs for sentiment classification \cite{Li:NAACL2016}.
We use the latter formulation of relevance and denote it as {\it Gradient}.
With this definition the relevance of an entire word is simply the squared $L_2$-norm of the prediction function's gradient w.r.t.\ the word embedding, i.e.\  $ R_{\boldsymbol{x_t}} = \|{\nabla}_{\boldsymbol x_t} \; f_c({\mathbf x})\|{_2^2}$\,.

A slight variation of this approach uses partial derivatives multiplied by the variable's value, i.e.\ $R_{x_i} =  \tfrac{\partial {f_c}}{\partial x_{i}}(\mathbf{x}) \cdot x_i$.
Hence, the word relevance is a dot product between prediction function gradient and word embedding: $ R_{\boldsymbol{x_t}} = ({\nabla}_{\boldsymbol x_t} \; f_c({\mathbf x}) )^T \, \boldsymbol{x_t}$\, \cite{Denil:2015}.
We refer to this variant as {\it Gradient$\times$Input}.

Both variants are general and can be applied to any neural network. They are computationally efficient and require one forward and backward pass through the net.

\subsection{Occlusion-based explanation}
Another method to assign relevances to single variables, or entire words, is by occluding them in the input, and tracking the difference in the network's prediction w.r.t.\ a prediction on the original unmodified input \cite{Zeiler:ECCV2014,Li:ArXiv2017}.
In computer vision the occlusion is performed by replacing an image region with a grey or zero-valued square \cite{Zeiler:ECCV2014}. In NLP word vectors, or single of its components, are replaced by zero; in the case of recurrent neural networks, the technique was applied to identify important words for sentiment analysis \cite{Li:ArXiv2017}.

Practically, the relevance can be computed in two ways: in terms of prediction function differences, or in the case of a classification problem, using a difference of probabilities,
i.e.\ $R_{x_i}=f_c( \mathbf{x} ) - f_c( \mathbf{x}_{| x_i=0} )$, or $R_{x_i}=P_c( \mathbf{x} ) - P_c( \mathbf{x}_{| x_i=0} )$,
where $P_c(\cdot)=  \tfrac {\exp f_c(\cdot)}{\sum_k{\exp f_k(\cdot)}}  $\,.
We refer to the former as {\it Occlusion}$_\text{f-diff}$, and to the latter as {\it Occlusion}$_\text{P-diff}$. Both variants can also be used to estimate the relevance of an entire word, in this case the corresponding word embedding is set to zero in the input.
This type of explanation is computationally expensive and requires $T$ forward passes through the network to determine one relevance value per word in the input sequence $\mathbf{x}$.

A slight variation of the above approach uses word omission (similarly to \citealp{Kadar:CL2017}) instead of occlusion. On a morphosyntactic agreement experiment (see \citealp{Poerner:ACL2018}), omission was shown to deliver inferior results, therefore we consider only occlusion-based relevance.

\subsection{Layer-wise relevance propagation}

A general method to determine input space relevances based on a backward decomposition of the neural network prediction function is layer-wise relevance propagation (LRP) \cite{Bach:PLOS2015}.
It was originally proposed to explain feed-forward neural networks such as convolutional neural networks \cite{Bach:PLOS2015,Bach:CVPR2016},
and was recently extended to recurrent neural networks \cite{Arras:WASSA2017,Ding:ACL2017,JAM:RUDDER2018}.

LRP consists in a standard forward pass, followed by a specific backward pass which is defined for each type of layer of a neural network by dedicated propagation rules.
Via this backward pass, each neuron in the network gets assigned a relevance, starting with the output neuron whose relevance is set to the prediction function's value, i.e.\ to $f_c(\mathbf{x})$.
Each LRP propagation rule redistributes iteratively, layer-by-layer, the relevance from higher-layer neurons to lower-layer neurons, and verifies a relevance conservation property\footnote{Methods based on a similar conservation principle include contribution propagation \cite{Landecker:CIDM2013}, Deep Taylor decomposition \cite{Montavon:PR2017}, and DeepLIFT \cite{Shrikumar:PMLR2017}.}. 
These rules were initially proposed in \citet{Bach:PLOS2015} and were subsequently justified by Deep Taylor decomposition \cite{Montavon:PR2017} for deep ReLU nets.

In practice, for a linear layer of the form $z_j = \sum_i z_i w_{ij} + b_j\;$, and given the relevances of the output neurons $R_j$,  
the input neurons' relevances $R_i$ are computed through the following summation:
$ R_{i} =  \sum_j \tfrac{z_i \cdot w_{ij}}{z_j \, + \, \epsilon \cdot {\text sign} (z_j)} \; \cdot R_j \,$,
where $\epsilon$ is a stabilizer (small positive number); this rule is commonly referred as $\epsilon$-LRP or $\epsilon$-rule\footnote{Such a rule was employed by previous works with recurrent neural networks \cite{Arras:WASSA2017,Ding:ACL2017,JAM:RUDDER2018},
although there exist also other LRP rules for linear layers (see e.g. \citealp{Montavon:DSP2018})}.
With this redistribution the relevance is conserved, up to the relevance assigned to the bias and ``absorbed'' by the stabilizer.

Further, for an element-wise nonlinear activation layer, the output neurons' relevances are redistributed identically onto the input neurons.

In addition to the above rules, in the case of a multiplicative layer of the form $z_j = z_g \cdot z_s$,
\citet{Arras:WASSA2017} 
proposed to redistribute zero relevance to the {\it gate} (the neuron that is sigmoid activated) i.e.\ $R_g=0$, and assign all the relevance to the remaining {\it signal} neuron (which is usually tanh activated) i.e.\ $R_s=R_j$. We call this LRP variant {\it LRP-all}, which stands for ``signal-take-all'' redistribution.
An alternative rule was proposed in \citet{Ding:ACL2017,JAM:RUDDER2018}, where the output neuron's relevance $R_j$ is redistributed onto the input neurons via:
$R_g = \tfrac {z_g } {z_g + z_s } R_j $ and $R_s = \tfrac {z_s } {z_g + z_s } R_j $. We refer to this variant as {\it LRP-prop}, for ``proportional'' redistribution.
We also consider two other variants.
The first one uses absolute values instead: $R_g = \tfrac {|z_g| } {|z_g| + |z_s| } R_j $ and $R_s = \tfrac {|z_s| } {|z_g| + |z_s| } R_j $, we call it  {\it LRP-abs}.
The second uses equal redistribution: $R_g = R_s = 0.5 \cdot R_j $ \cite{JAM:RUDDER2018}, we denote it as  {\it LRP-half}.
We further add a stabilizing term to the denominator of the {\it LRP-prop} and {\it LRP-abs} formulas, it has the form $\epsilon \cdot {\text sign} (z_g+z_s)$ in the first case, and simply $\epsilon$ in the latter.

Since the relevance can be computed in one forward and backward pass, the LRP method is efficient. Besides, it is general as it can be applied to any neural network made of the above layers:
it was applied successfully to CNNs, LSTMs, GRUs, and QRNNs \cite{Poerner:ACL2018,Yang:ICHI2018}\footnote{Note that in the present work we apply LRP to {\it standard} LSTMs, though \citet{JAM:RUDDER2018} showed that some LRP rules for product layers can benefit from simultaneously adapting the LSTM architecture.}.


\begin{table*}
	%
	\center
	\resizebox{\textwidth}{!}{%
	\begin{tabular}{lcccc}
		\toprule	
		Method                                                                           & Relevance  Formulation & Redistributed Quantity ($\sum_i R_{x_i}$) & Complexity    \\
		\midrule
		Gradient 			& $R_{x_i} = \big( \tfrac{\partial {f_c}}{\partial x_{i}}(\mathbf{x}) \big)^2$ &  $ \|{\nabla}_{\mathbf x} \; f_c({\mathbf x})\|{_2^2}$ & $\Theta(2 \cdot T)$\\
		Gradient$\times$Input 		& $R_{x_i} =  \tfrac{\partial {f_c}}{\partial x_{i}}(\mathbf{x}) \cdot x_i$ & $({\nabla}_{\mathbf x} \; f_c({\mathbf x}) )^T \, \mathbf{x}$ & $\Theta(2 \cdot T)$\\
		Occlusion 			& $R_{x_i}=f_c( \mathbf{x} ) - f_c( \mathbf{x}_{| x_i=0} )$ & - & $\Theta(T^{2})$\\
		LRP 				& backward decomposition of the neurons' relevance & $f_c({\mathbf x})$  & $\Theta(2 \cdot T)$ \\
		CD				& linearization of the activation functions & $f_c({\mathbf x})$ & $\Theta(T^{2})$\\		
		\bottomrule
	\end{tabular}%
	}

	\caption{Overview of the considered explanation methods. The last column indicates the computational complexity to obtain one relevance value per input vector, or word, where $T$ is the length of the input sequence.}

	\label{tab:methods_overview}
\end{table*}

\subsection{Contextual Decomposition}

Another method, specific to LSTMs, is contextual decomposition ({\it CD}) \cite{Murdoch:ICLR2018}. 
It is based on a {\it linearization} of the activation functions that enables to decompose the LSTM forward pass by distinguishing between two contributions: those made by a chosen contiguous {subsequence} (a word or a phrase) within the input sequence $\mathbf{x}$, and those made by the remaining part of the input. 
This decomposition results in a final hidden state vector $h_T$ (see the Appendix for a full specification of the LSTM architecture)
that can be rewritten as a sum of two vectors: $\beta_T$ and $\gamma_T$, where the former corresponds to the contribution from the ``relevant'' part of interest, and the latter stems from the ``irrelevant'' part.
When the LSTM is followed by a linear output layer of the form $w_c^T h_T + b_c\,$ for class $c$, then the relevance of a given word (or phrase) and for the {\it target} class $c$, is given by the dot product:
$w_c^T \beta_T$.

The method is computationally expensive as it requires $T$ forward passes through the LSTM to compute one relevance value per word.
Although it was recently extended to CNNs \cite{Murdoch:ICLR2019,Godin:EMNLP2018}, it is yet not clear how to compute the {\it CD} relevance in other recurrent architectures, or in networks with multi-modal inputs. 

See Table~\ref{tab:methods_overview} for an overview of the explanation methods considered in the present work.


\subsection{Methods not considered}

Other methods to compute relevances include Integrated Gradients \cite{Sundararajan:ICML2017}. It was previously compared against {\it CD} in \citet{Murdoch:ICLR2018}, and against the LRP variant of \citet{Arras:WASSA2017} in \citet{Poerner:ACL2018}, where in both cases it was shown to deliver inferior results.
Another method is DeepLIFT \cite{Shrikumar:PMLR2017}
, however, according to its authors, DeepLIFT was not designed for multiplicative connections, and its extension to recurrent networks remains an open question\footnote{Though \citet{Poerner:ACL2018} showed that, when using {only} the Rescale rule of DeepLIFT, and combining it with the product rule proposed in \citet{Arras:WASSA2017}, then the resulting explanations perform on-par with the LRP method of \citet{Arras:WASSA2017}}.
For a comparative study of explanation methods with a main focus on feed-forward nets, see \citet{Ancona:ICLR2018}\footnote{Note that in order to redistribute the relevance through product layers, \citet{Ancona:ICLR2018} simply relied on standard gradient backpropagation. Such a redistribution scheme is not appropriate for methods such as LRP, since it violates the relevance conservation property, hence their results for recurrent nets are not conclusive.}. For a broad evaluation of explanations, including several recurrent architectures, we refer to \citet{Poerner:ACL2018}. Note that the latter didn't include the {\it CD} method of \citet{Murdoch:ICLR2018}, and the LRP variant of \citet{Ding:ACL2017}, which we compare here.


\section{Evaluating Explanations}\label{section:evaluation}

\subsection{Previous work}
How to generally and objectively evaluate explanations, without resorting to {\it ad-hoc} evaluation procedures that are domain and task specific, is still active research \cite{Alishahi:2019,Belinkov:TACL2019}.

In computer vision, it has become common practice to conduct a perturbation analysis \cite{Bach:PLOS2015,Samek:TNNLS2017,Shrikumar:PMLR2017,Lundberg:NIPS2017,Ancona:ICLR2018,Chen:ICML2018,Morcos:ICLR2018}: hereby a few pixels in an image are perturbated (e.g.\ set to zero or blurred) according to their relevance (most relevant or least relevant pixels are perturbated first), and then the impact on the network's prediction is measured. 
The higher the impact, the more accurate is the relevance.

Other studies explored in which way relevances are consistent or helpful w.r.t.\ human judgment \cite{Ribeiro:KDD2016,Lundberg:NIPS2017,Nguyen:NAACL2018}. 
Some other works relied solely on the visual inspection of a few selected relevance heatmaps \cite{Li:NAACL2016,Sundararajan:ICML2017,Ding:ACL2017}.

In NLP, \citet{Murdoch:ICLR2018} proposed to measure the correlation between word relevances obtained on an LSTM, and the word importance scores obtained from a linear Bag-of-Words. However, the latter model ignores
the word ordering and context, which the LSTM can take into account, hence this type of evaluation is not adequate\footnote{The same way \citet{Murdoch:ICLR2018} try to ``match''  phrase-level relevances with n-gram linear classifier scores or human annotated phrases, but again this might be misleading, since the latter scores or annotations ignore the whole sentence context.}.
Other evaluations in NLP are task specific. For example \citet{Poerner:ACL2018} use the subject-verb agreement task proposed by \citet{Linzen:TACL2016}, where the goal is to predict a verb's number, and use the relevances to verify that the most relevant word is indeed the correct subject (or a noun with the predicted number).

Other studies include an evaluation on a synthetic task: \citet{Yang:ICHI2018} generated random sequences of MNIST digits and train an LSTM to predict if a sequence contains zero digits or not,
and verify that the explanation indeed assigns a high relevance to the zero digits' positions.

A further approach uses randomization of the model weights and data as sanity checks \cite{Adebayo:NIPS2018} to verify that the explanations are indeed dependent on the model and data.
Lastly, some evaluations are ``indirect'' and use relevances to solve a broader task, e.g.\ to build document-level representations \cite{Arras:PLOSONE2017}, or to redistribute predicted rewards in reinforcement learning \cite{JAM:RUDDER2018}.

\begin{table*}
	\scriptsize
	\center
	\begin{tabular}{lcccc}
		\toprule		
		
		{}    				        & $\rho({R_{\boldsymbol{x_a}}}, {n_a})$ 	& $\rho({R_{\boldsymbol{x_b}}}, {n_b})$		& {\tiny $ E[\frac{\vert {R_{\boldsymbol{x_a}}} \vert + \vert {R_{\boldsymbol{x_b}}} \vert}{\sum_{t}{ \vert R_{\boldsymbol{x_t}} \vert }} ]$ }	& {\tiny $E[(({R_{\boldsymbol{x_a}}}+{R_{\boldsymbol{x_b}}})-y_{\text{pred}})^2] $} 	\\
		\rule{0pt}{2ex}{}			& (in \%)					& (in \%)		 			& (in \%) 													        			& (``MSE'')										\\
		\rule{0pt}{1ex}{Toy Task Addition} & \multicolumn{4}{c}{} \\ 
		\cmidrule(l){1-5} 
		Gradient$\times$Input 			& \textbf{99.960} (0.017)			& \textbf{99.954} (0.019)			& \textbf{99.68} (0.53)			&	\textbf{24.10}{\boldmath${^{-4}}$} (8.10$^{-4}$)	\\		
		Occlusion 				& \textbf{99.990} (0.004)			& \textbf{99.990} (0.004)			& \textbf{99.82} (0.27)			&	\textbf{20.10}{\boldmath$^{-5}$} (8.10$^{-5}$)		\\	 
		LRP-prop				& \hspace{1mm}0.785 (3.619)			& 10.111 (12.362) 				& 18.14 (4.23)\hspace{1mm}		&	1.3 (1.0)						\\	 
		LRP-abs					& \hspace{1mm}7.002 (6.224)			& 12.410 (17.440) 				& 18.01 (4.48)\hspace{1mm}		&	1.3 (1.0)						\\	 
		LRP-half				& 29.035 (9.478)				& 51.460 (19.939) 				& 54.09 (17.53)				&	1.1 (0.3)						\\	 
		LRP-all					& \textbf{99.995} (0.002)			& \textbf{99.995} (0.002)			& \textbf{99.95} (0.05)\hspace{1mm} 	&	\textbf{2.10}{\boldmath$^{-6}$} (4.10$^{-6}$) 		\\	 
		CD					& \textbf{99.997} (0.002)			& \textbf{99.997} (0.002)			& \textbf{99.92} (0.06)\hspace{1mm}	&	\hspace{1mm}\textbf{4.10}{\boldmath$^{-5}$} (12.10$^{-5}$)		\\ 
		\rule{0pt}{3ex}{Toy Task Subtraction} & \multicolumn{4}{c}{} \\ 
		\cmidrule(l){1-5} 
		Gradient$\times$Input 			& \textbf{97.9} (1.6)\hspace{1mm}		& \textbf{-98.8} (0.6)\hspace{1mm}		& \textbf{98.3}  (0.6)\hspace{1mm}	&	\textbf{6.10}{\boldmath$^{-4}$} (4.10$^{-4}$)		\\		
		Occlusion 				& 99.0 (2.0)\hspace{1mm}			& -69.0 (19.1)					& 25.4  (16.8) 				&	0.05 (0.08)\hspace{1mm}					\\	 
		LRP-prop				& 3.1 (4.8) 					& \hspace{1mm}-8.4 (18.9)			& 15.0  (2.4)\hspace{1mm}		&	0.04 (0.02)\hspace{1mm}					\\
		LRP-abs					& 1.2 (7.6) 					& -23.0 (11.1)					& 15.1  (1.6)\hspace{1mm}		&	0.04 (0.002)						\\
		LRP-half				& \hspace{1mm}7.7 (15.3) 			& -28.9 (6.4)\hspace{1mm}			& 42.3  (8.3)\hspace{1mm}		&	0.05 (0.06)\hspace{1mm}					\\
		LRP-all					& \textbf{98.5} (3.5) 				& \textbf{-99.3} (1.3)\hspace{1mm}		& \textbf{99.3}  (0.6)\hspace{1mm}	&	\textbf{8.10}{\boldmath$^{-4}$} (25.10$^{-4}$)		\\	 
		CD					& -25.9 (39.1)					& -50.0 (29.2)					& 49.4  (26.1) 				&	0.05 (0.05)\hspace{1mm}					\\	 
		\bottomrule
	\end{tabular}
	
	\caption{Statistics of the relevance w.r.t.\ the input numbers $n_a$ and $n_b$ and the predicted output $y_{\text{pred}}$, on toy arithmetic tasks. $\rho$ denotes the correlation and $E$ the mean. Each statistic is computed over 2500 test data points. Reported are the mean (and standard deviation in parenthesis) over 50 trained LSTM models.}

	\label{tab:toy_experiment}
\end{table*}


\subsection{Toy Arithmetic Task}

As a first evaluation, we ask the following question: if we add two numbers within an input sequence, can we recover from the relevance the true input values?
This amounts to consider the adding problem \cite{Hochreiter:NIPS1996}, which is typically used to test the long-range capabilities of recurrent models \cite{Martens:ICML2011,Le:Arxiv2015}.
We use it here to test the faithfulness of explanations.
To that end, we define a setup similar to \citet{Hochreiter:NIPS1996}, but without {\it explicit} markers to identify the sequence start and end, and the two numbers to be added. 
Our idea is that, in general, it is  not clear what the ground truth relevance for a marker should be,
and we want only the relevant numbers in the input sequence to get a non-zero relevance.
Hence, we represent the input sequence $\mathbf{x} = (\boldsymbol{x_1}, \boldsymbol{x_2},...,\boldsymbol{x_T})$ of length $T$, with two-dimensional input vectors as:
\begin{equation*}
(\begin{smallmatrix} 
0		& 	0	&	0 	&      n_a     	& 	0 		& 0	& 0	        & n_{b}	         & 0		& 0    &   0   \\
n_{1} 		& 	...	& 	n_{a-1}	&	0	&	 n_{a+1} 	& ...	& n_{b-1}	& 0        	 & n_{b+1} 	& ...  &  n_{T}  
\end{smallmatrix})%
\end{equation*}
where the non-zero entries $n_t$ are random real numbers, and the two relevant positions $a$ and $b$ are sampled uniformly among $\{1,...,T\}$ with $a<b$.

More specifically, we consider two tasks that can be solved by an LSTM model with a hidden layer of size {\it one} (followed by a linear output layer with no bias\footnote{We omit the output layer bias since all considered explanation methods ignore it in the relevance computation, and we want to explain the ``full'' prediction function's value.}): the {\it addition} of signed numbers ($n_t$ is sampled uniformly from $[-1,-0.5] \cup [0.5, 1.0]$)
and the {\it subtraction} of positive numbers  ($n_t$ is sampled uniformly from $[0.5, 1.0]$\footnote{We avoid small numbers by using 0.5 as a minimum magnitude only to simplify learning, since otherwise this would encourage the model weights to grow rapidly.}).
In the former case the target output $y$ is $n_a+n_b$, in the latter it is $n_a-n_b$. During training we minimize Mean Squared Error (MSE).
To ensure that train/val/test sets do not overlap we use 10000 sequences with lengths $T \in \{4,...,10\}$ for training, 2500 sequences with $T \in \{11,12\}$ for validation, and 2500 sequences with $T \in \{13,14\}$ as test set.
For each task we train 50 LSTMs with a validation MSE $<10^{-4}$, the resulting test MSE is $<10^{-4}$.

Then, given the model's predicted output $y_{\text{pred}}$, we compute one relevance value $R_{\boldsymbol{x_t}}$ per input vector $\boldsymbol{x_t}$ (for the occlusion method we compute only {\it Occlusion}$_\text{f-diff}$ since the task is a regression; we also don't report {\it Gradient} results since it performs poorly).
Finally, we track the correlation between the relevances and the two input numbers $n_a$ and $n_b$. We also track the portion of relevance assigned to the relevant time steps, compared to the relevance for all time steps.
Lastly, we calculate the ``MSE'' between the relevances for the relevant positions $a$ and $b$ and the model's output.
Our results are compiled in Table~\ref{tab:toy_experiment}.

Interestingly, we note that on the addition task several methods perform well and produce a relevance that correlates perfectly with the input numbers: {\it Gradient$\times$Input}, {\it Occlusion}, {\it LRP-all} and {\it CD} (they are highlighted in bold in the Table).
They further assign all the relevance to the time steps $a$ and $b$ and almost no relevance to the rest of the input; and present a relevance that sum up to the predicted output.
However, on subtraction, only {\it Gradient$\times$Input} and {\it LRP-all} present a correlation of near one with $n_a$, and of near minus one with $n_b$. Likewise these methods assign only relevance to the relevant positions, and redistribute the predicted output entirely onto these positions. 

The main difference between our addition and subtraction tasks, is that the former requires only summing up the first dimension of the input vectors and can be solved by a Bag-of-Words approach (i.e. by ignoring the ordering of the inputs), while our subtraction task is truly {\it sequential} and requires the LSTM model to remember which number arrived first, and which number arrived second, via exploiting the gating mechanism. 

Since in NLP several applications require the word {\it ordering} to be taken into account to accurately capture a sentence's meaning (e.g.\ in sentiment analysis or in machine translation),
our experiment, albeit being an abstract numerical task, is pertinent and can serve as a first sanity check to check whether the relevance can reflect the {\it ordering} and the {\it value} of the input vectors. 

Hence we view our toy task as a minimal and unambiguous test (which besides being sequential, also exhibits a linear input-output relationship)
that acts as a necessary (though not sufficient) requirement for a recurrent neural network explanation method to be trustworthy in a more complex setup, where the ground truth relevance is less clear.

For the {\it Occlusion} method, the unreliability is probably due to the fact that the neural network has always seen two ``relevant'' input numbers in the input sequence during training, and therefore gets confused when one of these inputs is missing at the time of the relevance computation (``out-of-sample'' effect).
For {\it CD}, the weakness may come from the saturation of the activations, in particular of the gates, which makes their linearization induced by the {\it CD} decomposition inaccurate.


\begin{table*}
	%
	\center
	\resizebox{\textwidth}{!}{%
	\begin{tabular}{l|cccccccccc}
	\toprule	
	
	Accuracy Change \footnotesize (in \%)           & random 	& Grad. 	& Grad.$\times$Input     & LRP-prop      & LRP-abs          & LRP-half 	        & LRP-all 	    &   CD	       & Occlusion$_\text{f-diff}$      &  Occlusion$_\text{P-diff}$       \\
	\cmidrule(l){1-11} 											        			
	{decreasing order \footnotesize (std$<$16)}  	& 0		&    35    	&   66   		 &       15       &      -1         &      -3            &      \bf 97      &    \bf 92        &           \bf 96               &   \bf  100                       \\
	{increasing order \footnotesize (std$<$5)}      & 0		&    -18    	&   31  	         &       11       &      -1         &       3            &      \bf 49      &    36            &           \bf 50               &   \bf 100                        \\
	\bottomrule
	\end{tabular}%
	}

	\caption{Average change in accuracy when removing up to 3 words per sentence, either in {\it decreasing} order of their relevance (starting with correctly classified sentences), or in {\it increasing} order of their relevance (starting with falsely classified sentences). In both cases, the relevance is computed with the {\it true} class as the {\it target} class. Results are reported {\it proportionally} to the changes for i) random removal (0\% change)  and  ii) removal based on {\it Occlusion}$_\text{P-diff}$ (100\% change). For all methods, the higher the reported value the better. We boldface those methods that perform on-par with the occlusion-based relevances.}

	\label{tab:sentiment_word_removal}
\end{table*}

\subsection{5-Class Sentiment Prediction}\label{section:sentiment}
As a sentiment analysis dataset, we use the Stanford Sentiment Treebank \cite{Socher:EMNLP2013} which contains labels (very negative~$--$, negative~$-$, neutral~$0$, positive~$+$, very positive~$++$) for resp.\ 8544/1101/2210 train/val/test sentences and their constituent phrases.
As a classifier we employ the bidirectional LSTM from \citet{Li:NAACL2016}\footnote{\url{https://github.com/jiweil/Visualizing-and-Understanding-Neural-Models-in-NLP}}, which achieves 82.9\% binary, resp. 46.3\% five-class, accuracy on full sentences. 

\textbf{Perturbation Experiment}. 
In order to evaluate the {\it selectivity} of word relevances, we perform a perturbation experiment  aka ``pixel-flipping`` in computer vision \cite{Bach:PLOS2015,Samek:TNNLS2017}, i.e.\ we remove words from the input sentences according to their relevance, and track the impact on the classification performance.
A similar experiment has been conducted in previous NLP studies \cite{arras:ACL2016,Nguyen:NAACL2018,Chen:ICML2018}; and besides, such type of experiment can be seen as the input space pendant of {\it ablation}, which is commonly used to identify ``relevant'' intermediate neurons, e.g.\ in \citet{Lakretz:NAACL2019}.
For our experiment we retain test sentences with a length $\geq$ 10 words (i.e.\ 1849 sentences), and remove 1, 2, and 3 words per sentence\footnote{In order to remove a word we simply discard it from the input sentence and concatenate the remaining parts. We also tried setting the word embeddings to zero, which gave us similar results.}, according to their relevance obtained on the original sentence with the {\it true} class as the {\it target} class.
Our results are reported in Table~\ref{tab:sentiment_word_removal}.
Note that we distinguish between sentences that are initially correctly classified, and those that are initially falsely classified by the LSTM model.
Further, in order to condense the ``ablation'' results in a single number per method, we compute the accuracy decrease (resp.\ increase) {\it proportionally} to two cases: i) random removal, and ii) removal according to {\it Occlusion}$_\text{P-diff}$. 
Our idea is that random removal is the least informative approach, while {\it Occlusion}$_\text{P-diff}$
is the most informative one, since the relevance for the latter is computed in a similar way to the perturbation experiment itself, i.e.\ by deleting words from the input and tracking the change in the classifier's prediction. 
Thus, with this normalization, we expect the accuracy change (in \%) to be mainly rescaled to the range $[0, 100]$.


\begin{figure*}[h!]
\centering
\includegraphics[width=0.66\textwidth]{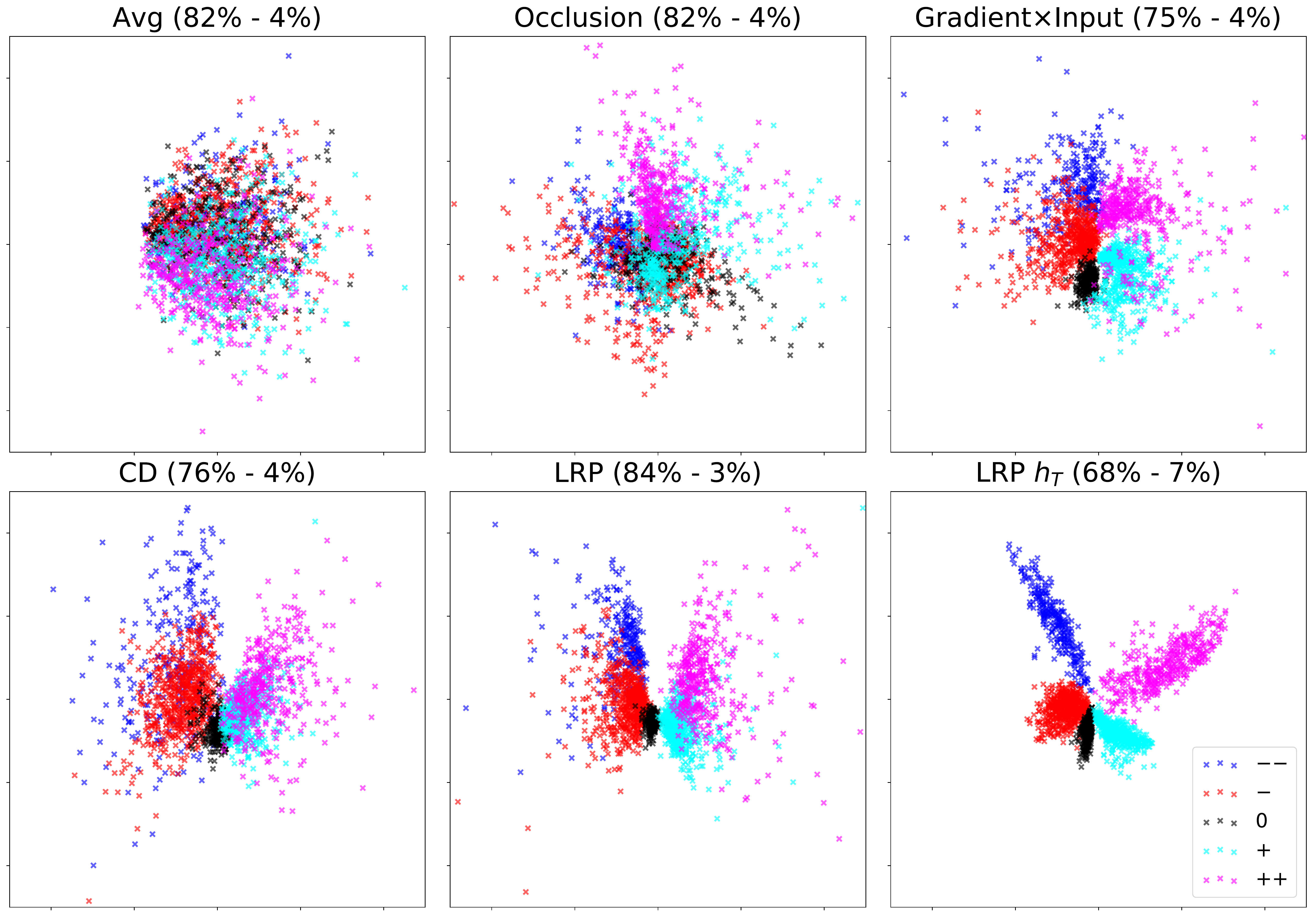}
\caption{PCA projection of sentence-level representations built on top of word embeddings that were linearly combined using their respective relevance. {\it Avg} corresponds to simple averaging of word embeddings.
For {LRP ${{h}_T}$} the last time step hidden layer was reweighted by its relevance. In parenthesis we indicate the percentage of variance explained by the first two PCA components (those that are plotted) and by the third PCA component.
The resulting representations were roughly ordered (row-wise) from less structured to more structured.}
\label{fig:PCA}
\end{figure*}

When removing words in decreasing order of their relevance, we observe that {\it LRP-all} and {\it CD} perform on-par with the occlusion-based relevance, with near 100\% accuracy change, followed by {\it Gradient$\times$Input} which performs only 66\%.  

When removing words in increasing order of their relevance (which mainly corresponds to remove words with a negative relevance), {\it Occlusion}$_\text{P-diff}$ performs best, followed by {\it Occlusion}$_\text{f-diff}$ and {\it LRP-all} (both around 50\%), then {\it CD} (36\%).
Unsurprisingly, {\it Gradient} performs worse than random, since its relevance is positive and thus low relevance is more likely to identify unimportant words for the classification (such as stop words), rather than identify words that {\it contradict} a decision, as noticed in \citet{Arras:WASSA2017}.
Lastly {\it Occlusion}$_\text{f-diff}$ is less informative than {\it Occlusion}$_\text{P-diff}$, since the former is not normalized by the classification scores for all classes. 

This analysis revealed that methods such as {\it LRP-all} and {\it CD} can detect influential words supporting (resp. contradicting) a specific classification decision, although they were not tuned towards the perturbation criterion, as opposed 
to {\it Occlusion} (which can be seen as the brute force approach to determine the inputs the model is the most sensitive to),
whereas gradient-based methods are less accurate in this respect.
Remarkably {\it LRP-all} only require one forward and backward pass to provide this information. 


\textbf{Sentence-Level Representations}.
In addition to testing selectivity, we explore if the word relevance can be leveraged to build sentence-level representations that present some regularities akin word2vec vectors.
For this purpose we linearly combine word embeddings using their respective relevance as weighting\footnote{W.l.o.g.\ we use here the {\it true} class as the {\it target} class, since the classifier's 5-class performance is very low. In a practical use-case one would use the {\it predicted} class instead.}.
For methods such as LRP and {\it Gradient$\times$Input} that deliver also relevances for {\it single} variables, we perform an element-wise weighting, i.e.\ we construct the sentence representation as:
$ \sum_t {R_{\boldsymbol{x_t}}} \odot {\boldsymbol{x_t}}$.
For every method we report the best performing variant from previous experiments, i.e.\ {\it Occlusion}$_\text{P-diff}$, {\it Gradient$\times$Input}, {\it CD} and {\it LRP-all}.
Additionally we report simple averaging of word embeddings (we call it {\it Avg}).
Further, for LRP, we consider an element-wise reweighting of the last time step hidden layer $h_T$ by its relevance, since LRP delivers also a relevance for each intermediate neuron (we call it LRP $h_T$).
We also tried using $h_T$ directly: this gave us a visualization similar to {\it Avg}.
The resulting 2D whitened PCA projections of the test set sentences are shown in Fig.~\ref{fig:PCA}.

Qualitatively LRP delivers the most structured representations, although for all methods the first two PCA components explain most of the data variance.
Intuitively it makes also sense that the neutral sentiment is located between the positive and negative sentiments, and that the very negative and very positive sentiments depart from their lower counterparts in the same vertical direction.

The advantage of having such regularities emerging via PCA projection, is that the sentence/phrase semantics might be investigated visually, without requiring any nonlinear dimensionality reduction like t-SNE (typically used to explore the representations learned by recurrent models, e.g. in \citealp{Cho:EMNLP2014,Li:NAACL2016}). Such representations might also be useful in information retrieval settings, 
where one could retrieve similar sentences/phrases by employing standard euclidean distance.


\begin{table*}
	%
	\center
	\resizebox{0.93\textwidth}{!}{%
	\begin{tabular}{ll|c|llll|r}
		\toprule	
		\multicolumn{2}{l|}{Composition}          	& Predicted 	& Heatmap  										&	\multicolumn{3}{l|}{Relevance}   				& \# samples\\
		\midrule
		1.& ``negated positive sentiment''		& $-$ 		& {\includegraphics[clip=true, trim=20mm 28.3cm 14.8cm 0.9cm]{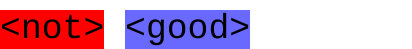}}	&	2.5$_{\;0.3}$ & -1.4$_{\;0.5}$ & {}				& 213	\\		
		2.& ``amplified positive sentiment''		& $++$ 		& {\includegraphics[clip=true, trim=20mm 28.3cm 14.8cm 0.9cm]{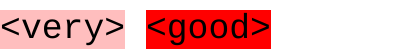}}	&	1.1$_{\;0.3}$ & 4.5$_{\;0.7}$ & {}				& 347	\\
		3.& ``amplified negative sentiment''		& $--$ 		& {\includegraphics[clip=true, trim=20mm 28.3cm 14.8cm 0.9cm]{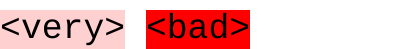}}	&	0.8$_{\;0.2}$ & 4.3$_{\;0.6}$ & {}				& 173	\\
		4.& ``negated amplified positive sentiment''	& $-$ 		& {\includegraphics[clip=true, trim=20mm 28.3cm 14.8cm 0.9cm]{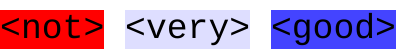}}	&	2.74$_{\;0.54}$ & -0.34$_{\;0.17}$ & -2.00$_{\;0.40}$		& 1745	\\
		\bottomrule
	\end{tabular}%
	}

	\caption{Typical heatmaps for various types of semantic compositions (indicated in first column), computed with the {\it LRP-all} method. The LSTM's {\it predicted} class (second column) is used as the {\it target} class. The remaining columns contain the average heatmap (positive relevance is mapped to red, negative to blue, and the color intensity is normalized to the maximum absolute relevance), the corresponding word relevance mean (and std as subscript), and the number of bigrams (resp.\ trigrams) considered for each type of composition.}

	\label{tab:semantic_composition}
\end{table*}


\begin{table}
	%
	\resizebox{0.92\columnwidth}{!}{%
	\begin{tabular}{l|c|l}
		\toprule	
		{N\textsuperscript{\underline{o}}}	& Predicted 	& Heatmap  												\\
		\midrule
		1 	& $--$ 		& {\includegraphics[clip=true, trim=20mm 28.3cm 10.9cm 0.9cm]{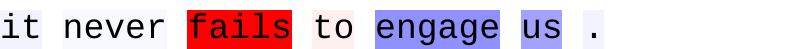}}	\\		
		1a	& $+$ 		& {\includegraphics[clip=true, trim=20mm 28.3cm 10.9cm 0.9cm]{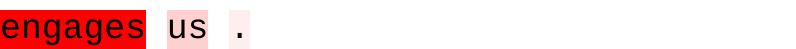}}		\\
		1b	& $-$ 		& {\includegraphics[clip=true, trim=20mm 28.3cm 10.9cm 0.9cm]{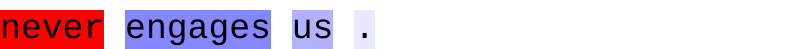}}		\\
		1c	& $--$ 		& {\includegraphics[clip=true, trim=20mm 28.3cm 10.9cm 0.9cm]{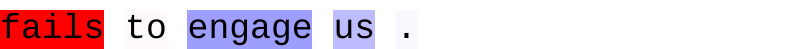}}		\\
		\midrule
		2 	& $++$ 		& {\includegraphics[clip=true, trim=20mm 28.3cm 10.9cm 0.9cm]{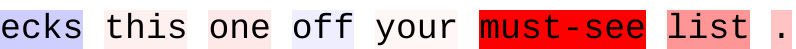}}	\\		
		2a	& $++$ 		& {\includegraphics[clip=true, trim=20mm 28.3cm 10.9cm 0.9cm]{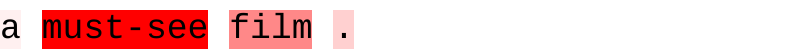}}		\\
		2b	& $--$ 		& {\includegraphics[clip=true, trim=20mm 28.3cm 10.9cm 0.9cm]{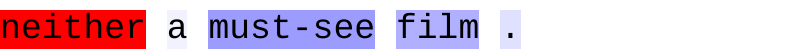}}		\\
		2c	& $++$ 		& {\includegraphics[clip=true, trim=20mm 28.3cm 10.9cm 0.9cm]{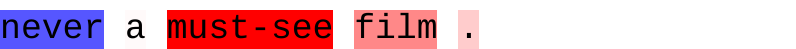}}		\\
		\bottomrule
	\end{tabular}%
	}

	\caption{Misclassified test sentences (1 and 2), and manually constructed sentences (1a-c, 2a-c). The LSTM's {\it predicted} class (second column) is used as the {\it target} class for the {\it LRP-all} heatmaps.}

	\label{tab:misclassified}
\end{table}

\section{Interpreting Single Predictions}\label{section:inspection}

Next, we analyze single predictions using the same task and model as in Section~\ref{section:sentiment}, and illustrate the usefulness of relevance visualization with {\it LRP-all}, which is the method that performed well in both our previous quantitative experiments.

\textbf{Semantic Composition}.
When dealing with real data, one typically has no ground truth relevance available. 
And the visual inspection of single relevance heatmaps can be counter-intuitive for two reasons:
the relevance is not accurately reflecting the reasons for the classifier's decision (the explanation method is bad), or the classifier made an error (the classifier doesn't work as expected).
In order to avoid the latter as much as possible, we automatically constructed bigram and trigram samples, which are built {\it solely} upon the classifier's predicted class, and visualize the resulting average relevance heatmaps
for different types of semantic compositions in Table~\ref{tab:semantic_composition}.
For more details on how these samples were constructed we refer to the Appendix, note though that in our heatmaps the negation \texttt{<not>}, the intensifier \texttt{<very>} and the sentiment words act as placeholders
for words with similar meanings, since the representative heatmaps were averaged over several samples.
In these heatmaps one can see that, to transform a positive sentiment into a negative one, the negation is predominantly colored as red, while the sentiment word is highlighted in blue, which intuitively makes sense since the explanation is computed {\it towards} the negative 
sentiment, and in this context the negation is responsible for the sentiment prediction.
For sentiment intensification, we note that the amplifier gets a relevance of the same sign as the amplified word, indicating the amplifier is {\it supporting} the prediction for the considered target class, but still  has less importance for the decision than
the sentiment word itself (deep red colored).
Both previous identified patterns also reflect consistently in the case of a negated amplified positive sentiment.


\textbf{Understanding Misclassifications}.
Lastly, we inspect heatmaps of misclassified sentences in Table~\ref{tab:misclassified}.
In sentence 1, according to the heatmap, the classifier didn't take the negation \texttt{never} into account, although it identified it correctly in sentence 1b. We postulate this is because of the strong sentiment
assigned to \texttt{fails} that overshadowed the effect of \texttt{never}.
In sentence 2, the classifier obviously couldn't grasp the meaning of the words preceding \texttt{must-see}.
If we use a negation instead, we note that it's taken into account in the case of \texttt{neither} (2b), but not in the case of \texttt{never} (2c),
which illustrates the complex dynamics involved in semantic composition, and that the classifier 
might also exhibit a bias towards the types of constructions it was trained on, which might then feel more ``probable'' or ``understandable'' to him.

Besides, during our experimentations, we empirically found that the {\it LRP-all} explanations are more helpful when using the classifier's {\it predicted} class as the {\it target} class (rather than the sample's {\it true} class), which intuitively makes sense since it's the class the model is the most confident about.
Therefore, to understand the classification of single samples, we generally recommend this setup.


\section{Conclusion}\label{section:conclusion}

In our experiments with standard LSTMs, we find that the LRP rule for multiplicative connections introduced in \citet{Arras:WASSA2017} performs consistently better than other recently proposed rules, such as the one from \citet{Ding:ACL2017}.
Further, our comparison using a 5-class sentiment prediction task highlighted that LRP is not equivalent to {\it Gradient$\times$Input} (as sometimes inaccurately stated in the literature, e.g.\ in \citealp{Shrikumar:PMLR2017}) and is more selective than the latter, which is consistent
with findings of \citet{Poerner:ACL2018}.
Indeed, the equivalence between {\it Gradient$\times$Input} and LRP holds only {if} using the $\epsilon$-rule with no stabilizer ($\epsilon=0$), and if the network contains {\it only} ReLU activations and max pooling as non-linearities \cite{Kindermans:arxiv2016,Shrikumar:arxiv2016}.
When using other LRP rules, or if the network contains other activations or product non-linearities (such as this is the case for LSTMs), then the equivalence does not hold (see \citet{Montavon:DSP2018} for a broader discussion).

Besides, we discovered that a few methods such as Occlusion \cite{Li:ArXiv2017} and CD \cite{Murdoch:ICLR2018} are not reliable and get inconsistent results on a simple toy task using an LSTM with only {\it one} hidden unit.

In the future, we expect decomposition-based methods such as LRP to be further useful to analyze character-level models, to explore the role of single word embedding dimensions, and to discover important hidden layer neurons.
Compared to attention weights (such as \citealp{Bahdanau:ICLR2015,Xu:ICML2015,OSMAN2019}), decomposition-based explanations take into account all intermediate layers of the neural network model, and can be related to a specific class.

\section*{Acknowledgments}
We thank Gr{\'e}goire Montavon for helpful discussions.
This work was supported by the German Federal Ministry for Education and Research through the Berlin Big Data Centre (01IS14013A), the Berlin Center for Machine Learning (01IS18037I) and the TraMeExCo project (01IS18056A). 
Partial funding by DFG is acknowledged (EXC 2046/1, project-ID: 390685689). This work was also supported by the Information \& Communications Technology Planning \& Evaluation (IITP) grant funded by the Korea government (No. 2017-0-00451).

\bibliographystyle{acl_natbib}

\appendix

\section{Appendix}
\label{sec:appendix}

\subsection{Long-Short Term Memory (LSTM) model}
All LSTMs used in the present work have the following recurrence form \cite{Hochreiter:1997, Gers:2000},
which is also the most commonly used in the literature \cite{Greff:15}:
\begin{equation*}
\begin{split}
  i_t &= \texttt{sigm} \;\; \Big( W_i \; h_{t-1} + U_i \; x_t + b_i \Big) \\
  f_t &= \texttt{sigm} \;   \Big( W_f \; h_{t-1} + U_f \; x_t + b_f \Big) \\
  o_t &= \texttt{sigm} \;   \Big( W_o \; h_{t-1} + U_o \; x_t + b_o \Big) \\
  g_t &= \texttt{tanh} \;   \Big( W_g \; h_{t-1} + U_g \; x_t + b_g \Big) \\
  c_t &= f_t \odot c_{t-1} \;  + \; i_t \odot g_t     \\
  h_t &= o_t \odot  \texttt{tanh} (c_t)
\end{split}
\end{equation*}
where $\mathbf{x} = (x_1, x_2,..., x_T)$  is the input sequence, $\texttt{sigm}$ and $\texttt{tanh}$ are element-wise activations, and $\odot$ is an element-wise multiplication.
The matrices $W$'s, $U$'s, and vectors $b$'s are connection weights and biases,
and the initial states $h_0$ and $c_0$ are set to zero. 
The resulting last time step hidden vector $h_T$ is ultimately fed to a fully-connected linear output layer yielding a prediction vector $\boldsymbol{f}(\mathbf{x})$, with one entry ${f_c}(\mathbf{x})$ per class.

The bidirectional LSTM \cite{Schuster:1997} we use for the sentiment prediction task, is a concatenation of two separate LSTM models as described above, each of them taking a different sequence of word embedding vectors as input. 
One LSTM takes as input the words in their original order, as they appear in the input sentence/phrase.
The other LSTM takes as input the same word sequence but in {\it reversed} order.
Each of these LSTMs yields a final hidden vector, say  $h^{\rightarrow}_T$ and $h^{\leftarrow}_T$. The concatenation of these two vectors is then fed to a fully-connected linear output layer, retrieving one prediction score ${f_c}(\mathbf{x})$ per class.

\subsection{Layer-wise Relevance Propagation (LRP) implementation}
We employ the code released by the authors \cite{Arras:WASSA2017} (\url{https://github.com/ArrasL/LRP_for_LSTM}), and adapt it to work with different LRP product rule variants.

In the toy task experiments, we didn't find it necessary to add any stabilizing term for numerical stability (therefore we use $\epsilon=0$ for all LRP rules).
In the sentiment analysis experiments, we use $\epsilon=0.001$ (except for the {\it LRP-prop} variant where we use $\epsilon=0.2$, we tried the following values: [0.001, 0.01, 0.1, 0.2, 0.3, 0.4, 1.0] and took the lowest one to achieve numerical stability).

\subsection{Contextual Decomposition (CD) implementation}
We employ the code released by the authors \cite{Murdoch:ICLR2018} (\url{https://github.com/jamie-murdoch/ContextualDecomposition}), and adapt it to work with a bidirectional LSTM.
We also made a slight modification w.r.t.\ the author's latest available version (commit e6575aa from March 30, 2018).
In particular in file \texttt{sent\_util.py} we changed line 125 to: \texttt{if i$>=$start and i$<$stop}, to exclude the stop index, and call the function \texttt{CD} with the arguments \texttt{start=k} and \texttt{stop=k+1} to compute the relevance
of the \texttt{k}-th input vector, or word, in the input sequence. This consistently led to better results for the {\it CD} method in all our experiments.

\subsection{Toy task setup}
As an LSTM model we consider a unidirectional LSTM with a hidden layer of size {one} (i.e.\ with one memory cell $c_t$), followed by a linear output layer with no bias.
Since the input is two-dimensional, this results in an LSTM model with 17 learnable parameters.
The weights are randomly initialized with the uniform distribution $U(-1.0, 1.0)$, and biases are initialized to zero.
We train the model with Pytorch's LBFGS optimizer, with an initial learning rate of 0.002, for 1000 optimizer steps,
and reduce the learning rate by a factor of 0.95 if the error doesn't decrease within 10 steps.
We also clip the gradient norm to 5.0. With this setting around 1/2 of the trained models on addition and 1/3 of the models for subtraction converged to a good solution with a validation MSE $<10^{-4}$.

\subsection{Semantic composition: generation of representative samples}
In a first step, we build a list of words with a positive sentiment~($+$), resp.\ a negative sentiment~($-$), as identified by the bidirectional LSTM model.
To that end, we predict the class of each word contained in the model's vocabulary, and select for each sentiment a list of 50 words with the highest prediction scores.
This way we try to ensure that the considered sentiment words are clearly identified by the model as being from the positive sentiment~($+$), resp.\ the negative sentiment~($-$) class.

In a second step, we build a list of negations and amplifiers.
To that end, we start by considering the same lists of 39 negations and 69 amplifiers as in \citet{Strohm:IEEE2018},
from which we retain only those that are classified as neutral (class~$0$) by the LSTM model, which leaves us with a list of 8 negations and 29 amplifiers.
This way we discard modifiers that are {\it biased} towards a specific sentiment, since our goal is to analyze the compositional effect of modifiers.

Then, for each type of considered semantic composition (see Table~\ref{tab:semantic_composition}), we generate bigrams resp.\ trigrams by using the previously defined lists of modifiers and sentiment words.

For compositions of type 1 (``negation of positive sentiment''), we note that among the constructed bigrams 60\% are classified as negative~($-$) by the LSTM model, 26\% are predicted as neutral~($0$), and  for the remaining 14\%
of bigrams the negation is not identified correctly and the corresponding bigram is classified as positive~($+$).
In order to remove negations that are ambiguous to the classifier, we retain only those negations which in at least 40\% of the cases predict the bigram as negative. These negations are: ['neither', 'never', 'nobody', 'none', 'nor'].
Then we average the results over all bigrams classified as negative~($-$).

For compositions of type 2 and 3 we proceed similarly.
For type 2 compositions (``amplification of positive sentiment''), we note that 29\% of the constructed bigrams are classified as very positive~($++$), and for type 3 compositions (``amplification of negative sentiment''),
24\% are predicted as very negative~($--$), while the remaining bigrams are of the same class as the original sentiment word (thus the amplification is not identified by the classifier).
Here again we retain only unambiguous modifiers, which in at least 40\% of the cases amplified the corresponding sentiment. The resulting amplifiers are: ['completely', 'deeply', 'entirely', 'extremely', 'highly', 'insanely', 'purely', 'really', 'so', 'thoroughly', 'utterly', 'very']
for type 2 compositions; and ['completely', 'entirely', 'extremely', 'highly', 'really', 'thoroughly', 'utterly'] for type 3 compositions. Then we average the results over the corresponding bigrams which are predicted as very positive~($++$), resp.\ very negative~($--$).

For type 4 compositions (``negation of amplified positive sentiment''), we construct all possible trigrams with the initial lists of negations, amplifiers and positive sentiment words.
We keep for the final averaging of the results only those trigrams where both the effect of the amplifier, and of the negation are correctly identified by the LSTM model.
To this end we classify the corresponding bigram formed by combining the amplifier with the positive sentiment word, and keep the corresponding sample if this bigram is predicted as very positive ~($++$).
Then we average the results over trigrams predicted as negative~($-$) (this amounts to finally retain 1745 trigrams).

We also tried to investigate the following composition: ``negation of negative sentiment'', similarly to compositions of type 1. However, we found that only 1\% of the constructed bigrams are classified as neutral~($0$), and that the remaining bigrams are classified as negative~($-$) (81\%) or even 
very negative~($--$) (18\%). This means, in most cases, negating a negative sentiment doesn't change the classifier's prediction, i.e. the negation is not detected by the LSTM model. Therefore  we did not retain this type of composition for constructing representative heatmaps.
That the impact of negation is not symmetric across different sentiments was also observed in previous works \cite{Socher:EMNLP2013,Li:NAACL2016}, and is probably due to the fact that some type of semantic compositions are more frequent than others in the training data (and more generally, in natural language) \cite{Fraenkel:2008}.

\end{document}